\documentclass[runningheads]{llncs}
\pdfoutput=1
\usepackage{amsmath,amssymb} 
\usepackage{subcaption}
\usepackage{multirow}
\usepackage{bm}
\usepackage{booktabs}
\usepackage{array}
\usepackage{mathtools}
\usepackage{graphicx}
\usepackage{comment}
\usepackage{color}
\usepackage{algorithmicx}
\usepackage[noend]{algpseudocode}
\usepackage[normalem]{ulem}
\usepackage{epsfig}
\usepackage{algorithm}
\usepackage[noend]{algpseudocode}
\usepackage{xparse}

\DeclareDocumentCommand{\jingAlgo}{ m O{\quad \leftarrow \quad} }{%
	{\rlap{$#1$} \hphantom{text}$#2$}%
}

\newcommand{\vect}[1]{\mathbf{#1}}

\begin{document}
\pagestyle{headings}
\mainmatter
\def\ECCVSubNumber{4256}  

\nocite{*}

\title{Benefiting Deep Latent Variable Models via Learning the Prior and Removing Latent Regularization} 

\titlerunning{Benefiting Deep Latent Variable Models}
%
\author{Rogan Morrow \and
Wei-Chen Chiu}
%
%
\institute{National Chiao Tung University, Taiwan \\
\email{rogan.o.morrow@gmail.com} \ \ \ \ \email{walon@cs.nctu.edu.tw}}
\maketitle

\begin{abstract}
There exist many forms of deep latent variable models, such as the variational autoencoder and adversarial autoencoder. Regardless of the specific class of model, there exists an implicit consensus that the latent distribution should be regularized towards the prior, even in the case where the prior distribution is learned. Upon investigating the effect of latent regularization on image generation our results indicate that in the case where a sufficiently expressive prior is learned, latent regularization is not necessary and may in fact be harmful insofar as image quality is concerned. We additionally investigate the benefit of learned priors on two common problems in computer vision: latent variable disentanglement, and diversity in image-to-image translation.
\keywords{VAEs, AAEs, generative autoencoders, disentanglement}
\end{abstract}

\begin{figure}
\centering
\begin{subfigure}{0.255\textwidth}
        \includegraphics[width=\textwidth]{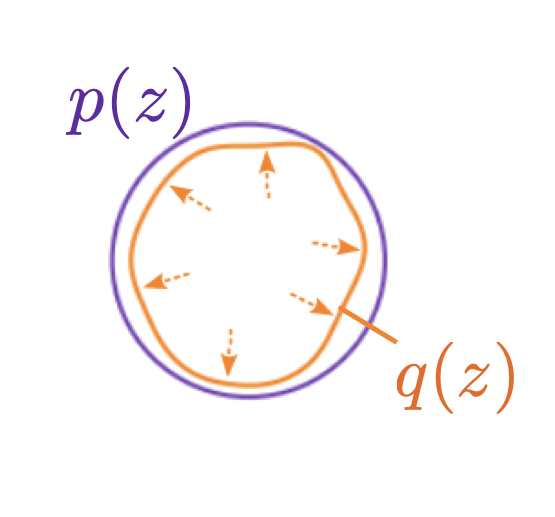}
        \caption{}
\end{subfigure} \hspace{0.275cm}
\begin{subfigure}{0.255\textwidth}
        \includegraphics[width=\textwidth]{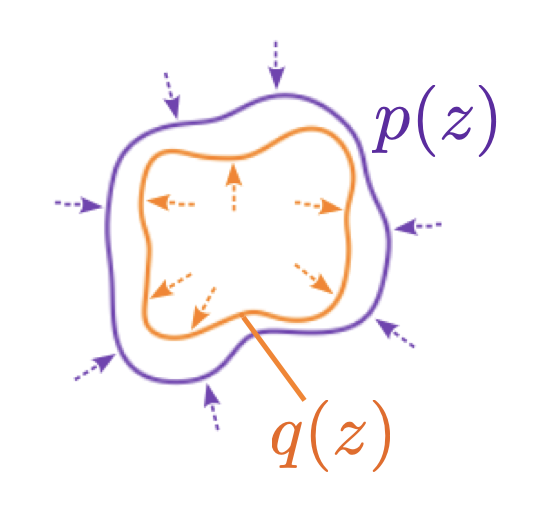}
        \caption{}
\end{subfigure} \hspace{0.25cm}
\begin{subfigure}{0.255\textwidth}
        \includegraphics[width=\textwidth]{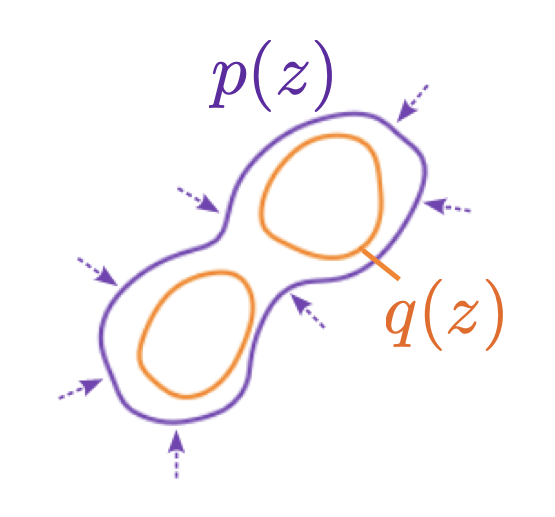}
        \caption{}
\end{subfigure}
\caption{Contour graph of prior distribution $p(\vect{z})$ and aggregated encoder distribution $q(\vect{z})$ for three different approaches to generative autoencoder training. Arrows represent forces acting on each distribution during training, excluding reconstruction loss. (a) Fixed $p(\vect{z})$, regularized $q(\vect{z})$. (b) Learned $p(\vect{z})$, regularized $q(\vect{z})$. (c) Learned $p(\vect{z})$, unregularized $q(\vect{z})$.}
\label{fig:learning_overview}
\end{figure}

\section{Introduction}
In the machine learning subfield of deep latent variable models, generative autoencoders such as variational autoencoders (VAEs) \cite{KingmaW13} and adversarial autoencoders (AAEs) \cite{MakhzaniSJG15} have attracted a significant amount of research interest \cite{BowmanVVDJB15,KingmaSW16,HuangLearnable2017,Rezende2018TamingV,Dai2018diagnosing,2017WAE_Tolstikhin}. Despite this, in their standard form they are still largely outperformed in terms of synthesized image quality by other deep generative models such as generative adversarial networks (GANs) \cite{2014GAN_Goodfellow,2017PGGAN,KarrasStyleGAN2018}, autoregressive models \cite{2016PixelCNN_Oord} and flow-based models \cite{DinhKB14,DinhSB16,KingmaGlow}. Even so, generative autoencoders maintain a number of properties that make them an attractive alternative, such as stable and efficient training as well as efficient synthesis.

In nearly all research done using such models, some form of regularization is imposed on the aggregated encoder distribution $q(\vect{z})$ in order to push it towards the prior distribution $p(\vect{z})$. For VAEs, this regularization exists in the form of a KL divergence between approximate posterior and prior, while AAEs force the aggregated posterior distribution to match the prior using an adversarial loss. This is of course necessary if the prior is fixed as is often the case, however we argue that when the prior is learnable it is possible to achieve a tight fit between aggregated posterior and prior without regularization, and that regularization in this case may actually have a negative impact on sample quality. Furthermore, removing regularization may result in a latent distribution that is beneficial to certain tasks such as disentanglement. Our contributions are as follows:
\begin{itemize}
\item We demonstrate empirically that when a sufficiently expressive prior $p(\vect{z})$ is learned, regularization of $q(\vect{z})$ is not necessary and may in fact be harmful to image quality.
\item We demonstrate that when the linear disentanglement metrics proposed in \cite{KarrasStyleGAN2018} are considered, a learned prior outperforms other methods commonly used for generative autoencoder disentanglement, and that regularization of $q(\vect{z})$ does not improve linear disentanglement. This is in contrast to the common method of adding stronger regularization for $q(z)$ such as in \cite{Matthey2017betaVAELB}.
\item We demonstrate that a learned prior is beneficial to sample diversity in multi-modal image-to-image translation tasks, where higher diversity is often a stated goal.
\end{itemize}

\section{Generative Autoencoders and Latent Regularization}
In this paper we focus on autoencoder-based generative models. This class of models defines an encoder distribution $q(\vect{z}|\vect{x})$ and a decoder distribution $p(\vect{x}|\vect{z})$, where the data $\vect{x} \sim p(\vect{x})$ is a random vector residing in space $\mathcal{X}$ and $\vect{z}$ is a latent code residing in space $\mathcal{Z}$. The negative log of $p(\vect{x}|\vect{z})$ is often referred to as the \textit{reconstruction loss}. Let $\vect{X} = \{\vect{x}^{(1)},...,\vect{x}^{(n)}\}$ be a set of i.i.d. observations drawn from the data distribution. Then the objective is to maximize
\begin{equation} \label{eq:1} \sum_{i=1}^n{\mathbb{E}_{\vect{z} \sim q(\vect{z}|\vect{x}^{(i)})}[\log p(\vect{x}^{(i)}|\vect{z})] - R(\vect{x}^{(i)})} \end{equation}
where $R(\vect{x})$ is a regularization term. Defining a prior $p(\vect{z})$ allows us to generate samples from the model by first sampling $\hat{\vect{z}} \sim p(\vect{z})$ and then sampling $\hat{\vect{x}} \sim p(\vect{x}|\hat{\vect{z}})$. Clearly in order for the generative distribution of the model to closely match $p(\vect{x})$, the aggregated encoder distribution $q(\vect{z}) = \int_{\mathcal{X}}{q(\vect{z}|\vect{x})p(\vect{x})}d\vect{x}$ should closely match the prior, such that we have $q(\vect{z}) \approx p(\vect{z})$ for all $\vect{z} \in \mathcal{Z}$. Therefore the regularization term $R(\vect{x})$ should be defined in such a way that it pushes $q(\vect{z})$ towards $p(\vect{z})$. Typically $p(\vect{z})$ is fixed as e.g. a standard normal distribution, however it is also possible to use the regularization term to learn the parameters of $p(\vect{z})$. If the prior is sufficiently expressive it may even be possible to remove regularization of $q(\vect{z})$ entirely so that the induced distribution of $q(\vect{z})$ is determined solely by pressure from the reconstruction loss, and the divergence between $q(\vect{z})$ and $p(\vect{z})$ is minimized solely by learning $p(\vect{z})$. These different approaches are visualized in Figure \ref{fig:learning_overview}. Examples of the first approach with fixed $p(\vect{z})$ are ubiquitous in the literature \cite{KingmaW13,KingmaSW16,MakhzaniSJG15,2017WAE_Tolstikhin}, indeed it would be possible to fill an entire page with references to previous works utilizing this approach. The second approach, with both learned $p(\vect{z})$ and regularized $q(\vect{z})$, is less common but still abundant \cite{ChenKSDDSSA16,TW:2017,2017arXiv170604223J}. The third approach with unregularized $q(\vect{z})$ is exceedingly rare however; the only previous works we are aware of that adopt this approach are \cite{LiSZ15,ma2018disentangled,pmlr-v80-bojanowski18a,2019GLF}, and other than \cite{2019GLF} they do not explicitly discuss the benefit of the approach\footnote{In \cite{2019GLF}, the authors focus specifically on VAE models and arrive at virtually the same result as us for our VAE derived model. However, their treatment of regularization and corresponding theoretical intepretion is fundamentally different to ours -- they experiment on the value of the inverse variance of the decoder distribution, and interpret their unregularized model as the vanishing noise limit of a VAE.}. This motivates the question: does such an approach have any benefits over the first two, and should it be more commonly used? Intuitively regularization should impede $q(\vect{z})$ from assuming a shape that is most beneficial to the decoder, and so one would assume that reconstruction loss would be negatively affected. We discuss other potential model-specific downsides in the following subsection.

\subsection{Related Models}
\noindent \textbf{Variational Autoencoders} \cite{KingmaW13} \\
The goal of a likelihood-based model is to maximize the likelihood
\begin{equation} \label{eq:2} p(\vect{x}^{(i)}) = \int_{\mathcal{Z}}{p(\vect{x}^{(i)}|\vect{z})p(\vect{z})d\vect{z}} \end{equation}
The integral in Eq. \ref{eq:2} is typically intractable, however. Variational autoencoders circumvent the issue of intractability by optimizing the variational lower bound
\begin{equation}\mathbb{E}_{\vect{z} \sim q(\vect{z}|\vect{x}^{(i)})}[\log p(\vect{x}^{(i)}|\vect{z})] - D_{KL}[q(\vect{z}|\vect{x}^{(i)})||p(\vect{z})] \leq \log p(\vect{x}^{(i)})\end{equation}
Thus we have $R(\vect{x}^{(i)}) = D_{KL}[q(\vect{z}|\vect{x}^{(i)})||p(\vect{z})]$. It is a well known issue that VAEs tend to not make full use of the latent code, as the objective becomes trapped in a local minima in which the posterior is close to the prior, a phenomenon known as ``posterior collapse'' \cite{he2018lagging}. Such a state occurs early on when the signal from the latent code is weak, resulting in a weak reconstruction term that is easily outweighed by the KL divergence term. This causes the posterior distributions of data points to overlap such that the optimal decoding becomes a weighted mean of the data points in pixel space, typically resulting in blurry reconstructions. This issue is particularly pernicious in the conditional setting if care is not taken, as it is easily possible for the model to entirely ignore the latent code when it is conditioned on a relevant context, resulting in a deterministic mapping.

Many methods for encouraging use of the latent code have been proposed. For instance, annealing the KL divergence term from $0$ to full strength \cite{BowmanVVDJB15,CWHuang17} allows the model to largely ignore the KL divergence term at the beginning of training. ``Free bits'', introduced in \cite{KingmaSW16}, places a limit on the information in nats per latent subset that can contribute to the KL divergence term, ensuring that each subset can contribute at least $\lambda$ nats of information without penalty. In the context of conditional variational autoencoders, \cite{ZhuBicycleGAN2017} proposed to add a latent reconstruction term to the objective to encourage the model to make full use of the latent code. All of these techniques are intended as a means of alleviating over-regularization imposed by the KL divergence term in the objective. If the prior is learned, however, it may be possible to eliminate such regularization entirely, hence obviating the need for any aforementioned techniques. \\

\noindent\textbf{Adversarial Autoencoders} \cite{MakhzaniSJG15} \\
Adversarial training \cite{2014GAN_Goodfellow} allows a distribution to be learned by playing a min-max game between a generator and a discriminator. The generator produces fake samples with the goal of fooling the discriminator, and the discriminator attempts to accurately classify samples as either real or fake. In the originally proposed setting where the discriminator outputs a probability, at optimality the model minimizes the Jensen-Shannon divergence between the data and generative distributions. Adversarial autoencoders apply this idea by using an adversarial term as the regularizer in Eq. \ref{eq:1}. The discriminator is trained separately to maximize
\begin{equation} \label{eq:4} \mathbb{E}_{\vect{z} \sim p(\vect{z})}[\log D(\vect{z})] + \mathbb{E}_{\vect{z} \sim q(\vect{z}|\vect{x}^{(i)})}[\log (1 - D(\vect{z})] \end{equation}
where $D$ is the discriminator network. Note that $R(\vect{x}^{(i)})$ is equal to Eq. \ref{eq:4}.

Regularization of the autoencoder in this way may introduce substantial noise. The reasoning for this is that discriminators are known to constantly shift their probability mass around during training in response to the generator, and so the decoder will be forced to deal with noisy latent codes. Removing the adversarial term from the autoencoder objective and instead using it to learn the prior dispels any such noise injection.

\section{The Unregularized Generative Autoencoder Objective and its Connections to Optimal Transport}
Removal of the regularization term in Eq. \ref{eq:1} presents an immediate problem -- we desire to rely solely on the learning of $p(\vect{z})$ to minimize the divergence between $p(\vect{z})$ and $q(\vect{z})$, however $p(\vect{z})$ relies on the regularization term to learn. Therefore we must reformulate the objective so that learning of the prior is possible while the latent distribution remains unregularized. This can be achieved by casting the objective as a bilevel optimization problem. Let the encoder, decoder and prior distributions be denoted by $q_{\phi}(\vect{z}|\vect{x})$, $p_{\psi}(\vect{x}|\vect{z})$ and $p_{\theta}(\vect{z})$, and parameterized by $\phi$, $\psi$ and $\theta$ respectively. Then the objective becomes
\begin{subequations}
\label{eq:bilevel}
\begin{align}
\max_{\phi, \psi} &\ F(\phi, \psi, \theta) \\ 
\text{s.t.  } & \theta \in \arg \max_{\theta} f(\phi, \theta)
\end{align}
\end{subequations}
where $F$ and $f$ are the upper and lower-level objectives and are given by
\begin{align}
\label{eq:upperlevel}
F(\phi, \psi, \theta) &= \sum_{i=1}^n{\mathbb{E}_{\vect{z} \sim q_{\phi}(\vect{z}|\vect{x}^{(i)})}[\log p_{\psi}(\vect{x}^{(i)}|\vect{z})] - \beta R(\vect{x}^{(i)};\phi,\theta)} \\
f(\phi, \theta) &= \sum_{i=1}^n{-R(\vect{x}^{(i)};\phi,\theta)}
\end{align}
where we have introduced a hyperparameter $\beta$ to control the strength of the regularization term; setting $\beta = 0$ allows us to remove regularization entirely without affecting learning of the prior. Note that when $\beta = 1$ the objective is equivalent to Eq. \ref{eq:1}, therefore our approach is consistent with the original objective. The problem in Eq. \ref{eq:bilevel} can be optimized straightforwardly via simultaneous gradient ascent by updating $\phi$, $\psi$ and $\theta$ using $\frac{\partial F}{\partial \phi}$, $\frac{\partial F}{\partial \psi}$ and $\frac{\partial f}{\partial \theta}$ respectively~\cite{STN2019}.
Consider the case where $\beta = 0$ and the summation over $R(\vect{x})$ in $f$ corresponds to a divergence measure. If $p_{\theta}(\vect{z})$ is expressive enough to match any induced $q_{\phi}(\vect{z})$, then Eq. \ref{eq:bilevel} is equivalent to the objective in the main theorem of \cite{2017WAE_Tolstikhin}, in which the authors demonstrate the equivalence between the optimal transport objective and
\begin{subequations}
\label{eq:ot}
\begin{align}
\min_{\phi,\psi}\ & \mathbb{E}_{\vect{x} \sim p(\vect{x})}[\mathbb{E}_{\vect{z} \sim q_{\phi}(\vect{z}|\vect{x})}[c(\vect{x}, G_{\psi}(\vect{z}))]] \\
\text{s.t. } & \mathcal{D}_{\vect{z}}(q_{\phi}(\vect{z}) || p_{\theta}(\vect{z})) = 0
\end{align}
\end{subequations}
where $\mathcal{D}_{\vect{z}}$ is an arbitrary divergence measure, $c$ is a cost function and $G_{\psi}$ is a deterministic mapping $\mathcal{Z} \rightarrow \mathcal{X}$. The cost function is defined implicitly in Eq. \ref{eq:upperlevel} through the reconstruction loss, for example
\begin{equation}
p_{\psi}(\vect{x}|\vect{z}) = \mathcal{N}(\vect{x}|G_{\psi}(\vect{z}),\bm{I})
\end{equation}
gives us the L2 cost function
\begin{equation}
c(\vect{x}, G_{\psi}(\vect{z})) = \lVert \vect{x} - G_{\psi}(\vect{z}) \rVert^2_2
\end{equation}
plus a normalizing constant, in which case the model is performing 2-Wasserstein distance minimization between the data and generative distributions. When sampling from the generative distribution, in order to be consistent with the optimal transport interpretation it becomes necessary to use the deterministic mapping $G_{\psi}(\vect{z})$ rather than sampling from the full decoder distribution $p_{\psi}(\vect{x}|\vect{z})$, however this is already common practice when using decoders with simple distributions such as isotropic Gaussians. In practice, $p_{\theta}(\vect{z})$ will typically not be expressive enough to exactly match any induced $q_{\phi}(\vect{z})$, in which case Eq. \ref{eq:bilevel} can be viewed as a relaxation of the constraint in Eq. \ref{eq:ot}. In \cite{2017WAE_Tolstikhin} the authors also propose a relaxed objective by removing the constraint and adding a penalty to the objective, which coincides with Eq. \ref{eq:1}. Note that when $R(\vect{x})$ is derived from the AAE objective, $\mathcal{D}_{\vect{z}}$ corresponds to the Jensen-Shannon divergence, whereas when $R(\vect{x})$ is derived from the VAE objective, $\mathcal{D}_{\vect{z}}$ corresponds to the Kullback-Leibler divergence.

\section{Effect of Regularization: An Experimental Setup}
Our goal here is to answer the question: does regularization of $q(\vect{z})$ hurt or help image quality when the prior is learned? We could simply compare two models, one with full regularization and one without any regularization, however it is possible that image quality as a function of regularization strength is not monotonic, and so we investigate how image quality varies with the strength of regularization by running experiments across varying values of $\beta$.
We consider two kinds of models in our approach: VAEs with prior learned via normalizing flow, and AAEs with prior learned by a simple MLP. Normalizing flows \cite{2015NormalizingFlows_Rezende} optimize a composition of bijective functions $f = f_T \circ ... \circ f_1$ using the change of variables formula $p(\vect{u}) = p(\vect{h})|\det (\frac{\partial \vect{h}}{\partial \vect{u}^T})|$, where $\vect{h} = f(\vect{u})$ and $p(\vect{h})$ is typically a standard normal distribution. Using a normalizing flow to learn the prior of a VAE was first proposed in \cite{ChenKSDDSSA16}, where the authors use an autoregressive flow. In our experiments we use a flow with affine coupling layers as proposed in \cite{DinhSB16}. This gives rise to a set of latents $\{\vect{z}_t \in \mathcal{Z}_t\}_{t=0}^T$, where $\vect{z}_T \sim q_{\phi}(\vect{z})$ and $\vect{z}_0$ should approximate $\mathcal{N}(\vect{0},\bm{I})$ after training. In \cite{ChenKSDDSSA16} the authors optimize a single objective
\begin{equation}\sum_{i=1}^n{\mathbb{E}_{\vect{z} \sim q_{\phi}(\vect{z}|\vect{x}^{(i)})}[\log p_{\psi}(\vect{x}^{(i)}|\vect{z}) + \beta(\log p_{\theta}(\vect{z}) - \log q_{\phi}(\vect{z}|\vect{x}^{(i)}))]} \end{equation}
Reformulating the objective so that it conforms to the bilevel structure of Eq. \ref{eq:bilevel} we have
\begin{align} F(\phi, \psi, \theta) &= \sum_{i=1}^n{\mathbb{E}_{\vect{z} \sim q_{\phi}(\vect{z}|\vect{x}^{(i)})}[\log p_{\psi}(\vect{x}^{(i)}|\vect{z}) + \beta(\log p_{\theta}(\vect{z}) - \log q_{\phi}(\vect{z}|\vect{x}^{(i)}))]} \\
\label{eq:6} f(\phi, \theta) &= \sum_{i=1}^n{\mathbb{E}_{\vect{z} \sim q_{\phi}(\vect{z}|\vect{x}^{(i)})}[\log p_{\theta}(\vect{z})]} \end{align}
For VAEs with a fixed prior (e.g. a standard normal distribution), as $\beta$ becomes larger we achieve a more structured latent space at the expense of reconstruction quality \cite{Matthey2017betaVAELB}, and vice versa as $\beta$ becomes smaller. When the parameters of the prior $p_{\theta}(\vect{z})$ are learnable, however, decreasing $\beta$ does not necessarily sacrifice latent structure, as any additional incurred divergence between the aggregate posterior $q_{\phi}(\vect{z})$ and the prior $p_{\theta}(\vect{z})$ can be mitigated by adjusting the parameters of $p_{\theta}(\vect{z})$. In our experiments we consider VAEs with reconstruction loss given by an isotropic Gaussian, i.e. $p_{\psi}(\vect{x}|\vect{z}) = \mathcal{N}(\vect{x}|G_{\psi}(\vect{z}),\gamma \bm{I})$ where $G_{\psi}(\vect{z})$ is the output of the decoder. It is common in many VAE implementations to keep the variance fixed, i.e. $\gamma$ is fixed to a predetermined value and not altered during training. In \cite{Dai2018diagnosing} the authors propose learning the variance of the decoder distribution,  and prove that it is always possible to achieve a better VAE cost by lowering the value of $\gamma$. We therefore consider both fixed $\gamma = 1$ and learned $\gamma$ approaches in our experiments. Note that $\gamma$ in effect, similarly to $\beta$, changes the strength of the regularization term. As $\gamma$ becomes smaller the decoder distribution becomes more peaked, and so when the decoder has a good estimate of the mean the reconstruction loss will be much stronger relative to the KL divergence term. When $\gamma$ is learned, the decoder is incentivized to lower the value of $\gamma$ whenever it obtains a better estimate of the mean, and so we can expect that $\beta$ will have less effect on the sample quality of the model than when $\gamma$ is fixed. Learning the prior becomes necessary in this case, however, as the model will prioritize learning the data manifold over learning the ground truth distribution as pointed out in \cite{Dai2018diagnosing}.

\begin{figure}[t]
\centering
\begin{subfigure}{0.35\textwidth}
        \includegraphics[width=\textwidth]{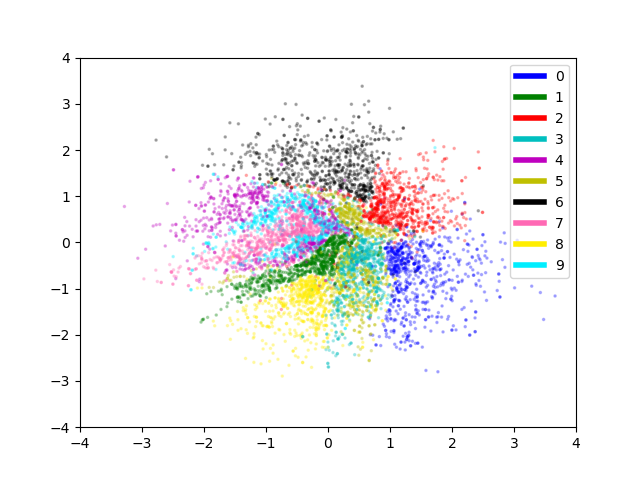}
        \caption{}
\end{subfigure}
\begin{subfigure}{0.35\textwidth}
        \includegraphics[width=\textwidth]{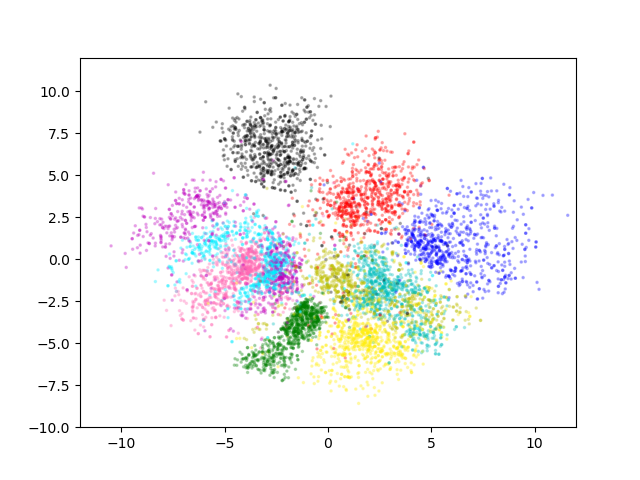}
        \caption{}
\end{subfigure}
\begin{subfigure}{0.35\textwidth}
        \includegraphics[width=\textwidth]{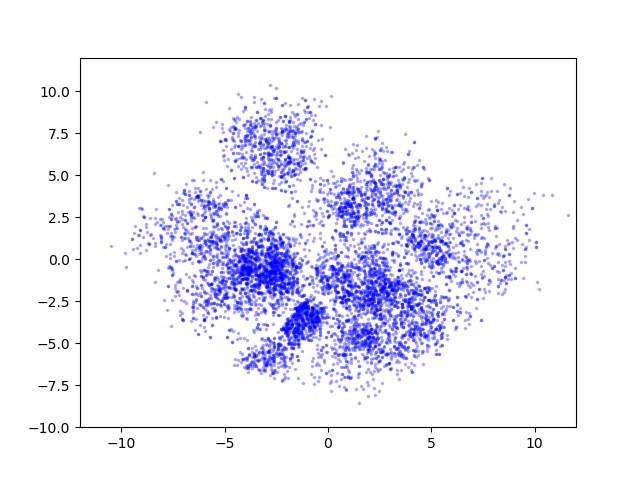}
        \caption{}
\end{subfigure}
\begin{subfigure}{0.35\textwidth}
        \includegraphics[width=\textwidth]{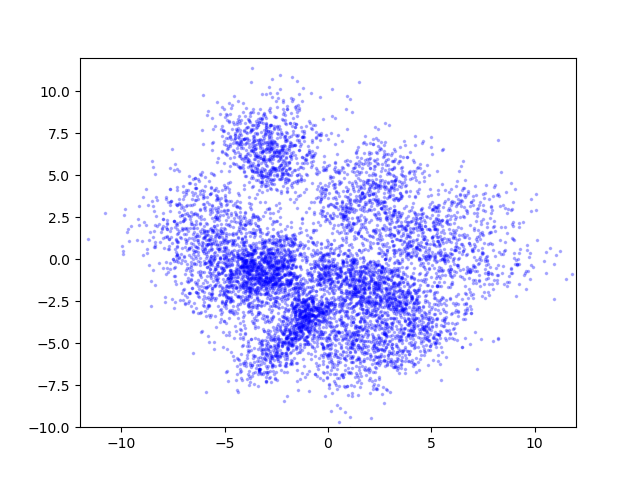}
        \caption{}
\end{subfigure}
\caption{Distribution of the latents of a VAE with flow prior with $\beta = 0$ after training on MNIST with $\dim(\mathcal{Z}) = 2$. (a) Distribution of $f(\vect{z})|_{\vect{z} \sim q(\vect{z})}$. Datapoints are colored according to class. (b) Distribution of $q(\vect{z})$. (c) Same as (b) but without class coloring. (d) Distribution of $f^{-1}(\vect{z})|_{\vect{z} \sim \mathcal{N}(\vect{0}, \bm{I})}$. It can be seen that while the autoencoder learns a complex latent distribution with classes well separated, the normalizing flow is able to learn a close match.}
\label{fig:z_latent_distribution}
\end{figure}

AAEs with learnable priors were first proposed in \cite{2017arXiv170604223J}. As we did for the VAE model, we again split the objective in order to conform with the bilevel objective in Eq. \ref{eq:bilevel}. Formally, we have the following objectives:
\begin{align}
F(\phi, \psi, \theta) &= \sum_{i=1}^n{\mathbb{E}_{\vect{z} \sim q_{\phi}(\vect{z}|\vect{x}^{(i)})}[\log p_{\psi}(\vect{x}^{(i)}|\vect{z})  - \beta \log (1 - D_{\omega}(\vect{z}))]} \\
f(\phi, \theta) &= \sum_{i=1}^n{\mathbb{E}_{\vect{z} \sim p_{\theta}(\vect{z})}[- \log D_{\omega}(\vect{z}))]} \\
g(\omega, \phi, \theta) &= \sum_{i=1}^n{\mathbb{E}_{\vect{z} \sim p_{\theta}(\vect{z})}[\log D_{\omega}(\vect{z})] + \mathbb{E}_{\vect{z} \sim q_{\phi}(\vect{z}|\vect{x}^{(i)})}[\log (1 - D_{\omega}(\vect{z}))]}
\end{align}
where $D_{\omega}$ denotes the discriminator and $\omega$ denotes the parameters of the discriminator network. The discriminator objective is represented by $g$, and is optimized by $\arg \max_{\omega} g(\omega, \phi, \theta)$. These three objectives can be maximized iteratively. Sampling from the prior is performed by first sampling $u \sim \mathcal{N}(\vect{0}, \bm{I})$ and then passing the sample through a simple MLP parameterized by $\theta$.

\section{Experiments on values of $\beta$}
\label{sec:mnist_cifar10}

\begin{figure*}[t]
\centering
\begin{subfigure}{0.4\textwidth}
        \includegraphics[width=\textwidth]{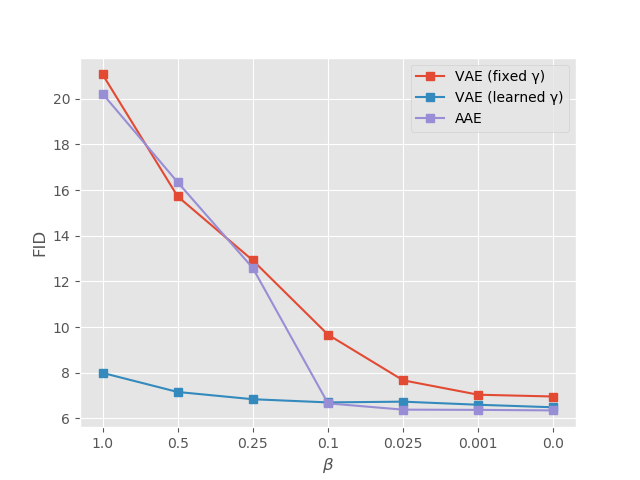}
        \caption{MNIST}
\end{subfigure}
\begin{subfigure}{0.4\textwidth}
        \includegraphics[width=\textwidth]{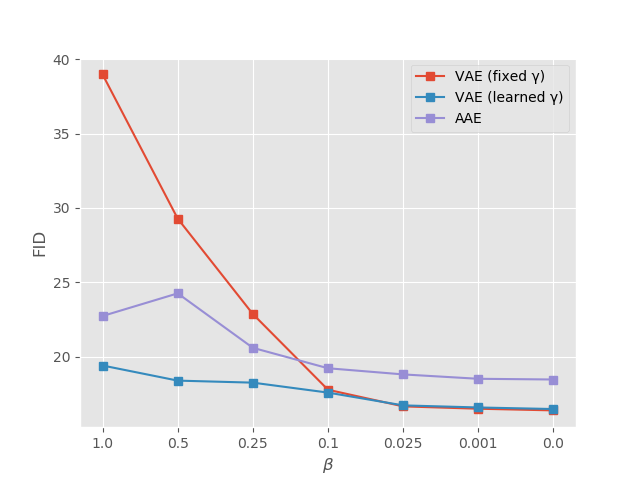}
        \caption{Fashion-MNIST}
\end{subfigure} \\
\begin{subfigure}{0.4\textwidth}
        \includegraphics[width=\textwidth]{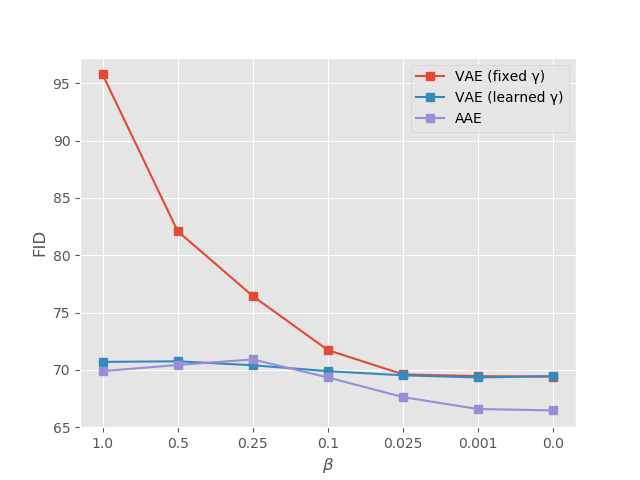}
        \caption{CIFAR-10}
\end{subfigure}
\begin{subfigure}{0.4\textwidth}
        \includegraphics[width=\textwidth]{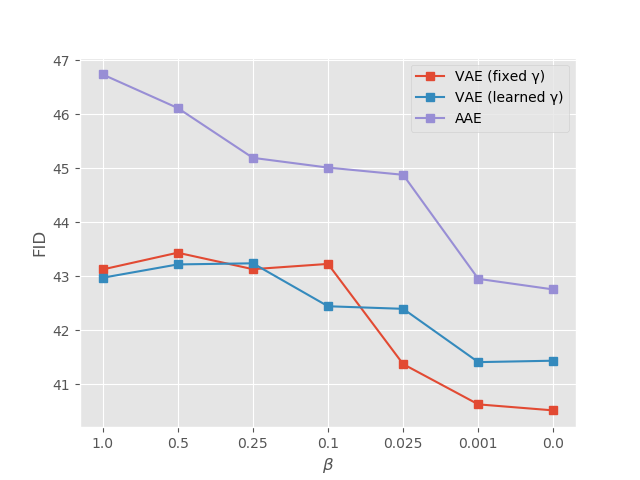}
        \caption{CelebA}
\end{subfigure}
\caption{Experimental results using different values of $\beta$ for different models. X-axis represents $\beta$, not to scale. Y-axis represents FID score.}
\label{fig:beta_results}
\end{figure*}

We first train on MNIST a VAE with normalizing flow prior and $\beta$ set to zero with a 2-dimensional latent code in order to demonstrate visually the learned latent distributions. This is shown in Figure \ref{fig:z_latent_distribution}; while the autoencoder has learned a latent distribution that is complex and multi-modal, the samples from the learned prior are a close match.

We experimented with varying values of $\beta$ on the MNIST, Fashion-MNIST, CIFAR-10 and CelebA datasets. We chose to adopt the Fr\'{e}chet Inception Distance (FID) \cite{HeuselRUNKH17} to measure image quality, a common measure used in GAN evaluation, for our quantitative comparisons. FID scores are given by the Fr\'{e}chet distance between layer activations of the Inception v3 network \cite{SzegedyVISW15}, with lower scores indicating greater similarity between two image sets.

\begin{figure}[t]
\centering
\begin{subfigure}{0.42\textwidth}
\includegraphics[width=\textwidth]{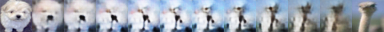}
\includegraphics[width=\textwidth]{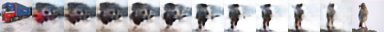}
\includegraphics[width=\textwidth]{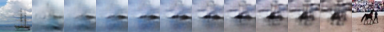}
\includegraphics[width=\textwidth]{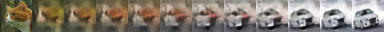}
\includegraphics[width=\textwidth]{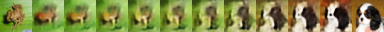}
\end{subfigure}
\begin{subfigure}{0.42\textwidth}
\includegraphics[width=\textwidth]{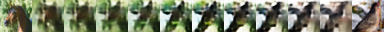}
\includegraphics[width=\textwidth]{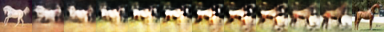}
\includegraphics[width=\textwidth]{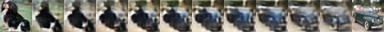}
\includegraphics[width=\textwidth]{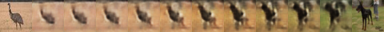}
\includegraphics[width=\textwidth]{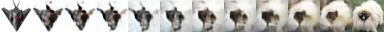}
\end{subfigure}
\caption{Interpolation between samples from the CIFAR-10 test set for a flow prior with fixed $\gamma$ and $\beta = 0$. Leftmost and rightmost columns contain real images from the test set before encoding, middle columns contain interpolations between them.}
\label{fig:cifar10_interpolation}
\end{figure}

Results are reported in Figure \ref{fig:beta_results}. We report the average across 5 runs, and random seeds were kept fixed between runs such that the only changing hyperparameter is $\beta$. It can be seen that FID scores typically decrease as $\beta$ is decreased for all models, suggesting that regularization of $q(z)$ is unnecessary and in fact potentially harmful to image quality. Interpolations in latent space for a VAE with L2 decoder and $\beta$ set to zero are shown in Figure \ref{fig:cifar10_interpolation} in order to demonstrate that the model has learned a smooth manifold. We use spherical linear interpolation as suggested by \cite{White16a}.

FID scores were calculated against test sets using 10,000 samples. When calculating FID we used exactly the same code as was used in \cite{Dai2018diagnosing}. Although there may be slight discrepancies between different implementations of FID score, we stress that our experiments are meant to be self-contained: we are primarily concerned with how $\beta$ affects the same model trained under the same conditions, rather than how our results compare with those in other papers. We used a latent dimensionality of 64 for all datasets, and the same network architecture as was used in \cite{ChenDHSSA16}. We note that we intentionally used priors that were of sufficient expressiveness to learn the latent distribution. For normalizing flows especially, this can result in a significant number of parameters being added to the model. We acknowledge that in cases where a lightweight model is desired or required such that a high capacity prior is not feasible, our results are not applicable, as latent regularization may still be required in order to achieve a good fit between encoder and prior distributions. For more details regarding the model architecture and training details, please see the supplementary.

\begin{figure*}[t]
\centering
\begin{subfigure}{0.4\textwidth}
        \includegraphics[width=\textwidth]{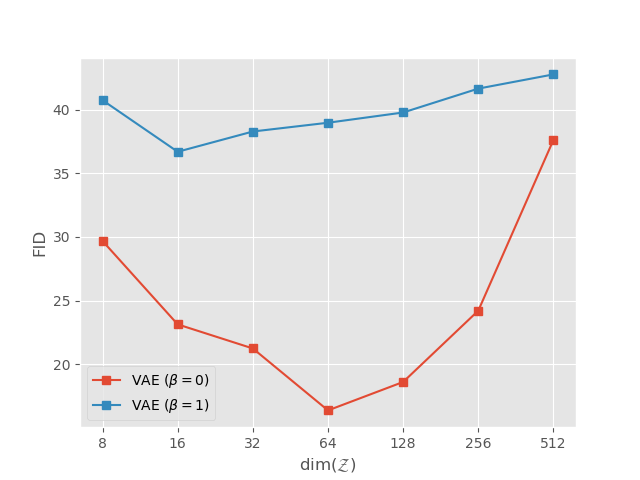}
        \caption{Fashion-MNIST}
\end{subfigure}
\begin{subfigure}{0.4\textwidth}
        \includegraphics[width=\textwidth]{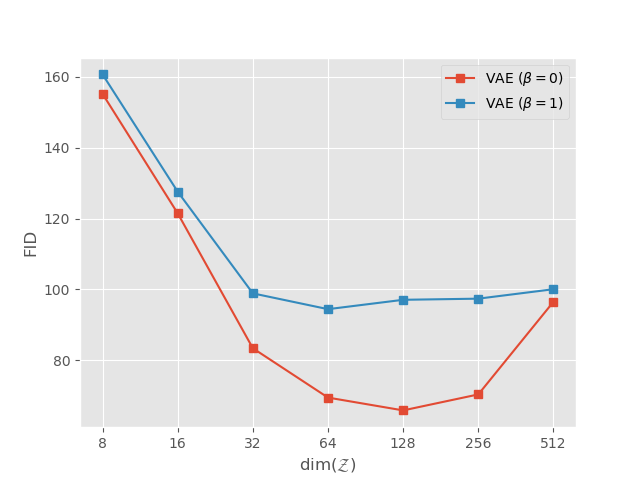}
        \caption{CIFAR-10}
\end{subfigure}
\caption{Experimental results using different values for the latent dimensionality. X-axis represents the latent dimensionality. Y-axis represents FID score.}
\label{fig:latent_dim_results}
\end{figure*}

Since there is no regularization imposed on the latent distribution at all when $\beta = 0$, it is possible that the dimensionality of the latent space becomes a critical hyperparameter when tuning the model. This is because it may be necessary to create an information bottleneck to induce a latent distribution that allows for the model to generalize well. An information bottleneck is also desirable for inducing a latent distribution that is easy enough for the prior network to learn. We experimented with how different values for the latent dimensions affects FID score for a VAE when $\beta = 0$, results are shown in Figure \ref{fig:latent_dim_results}. We also included results for when $\beta = 1$ for reference. Note that the exact same prior network settings from the previous experiments was used, which was tuned for a latent dimensionality of 64.
Results indicate that while lower values of $\beta$ may achieve better image quality, it may be necessary to carefully tune the prior network and latent dimensionality, whereas standard VAEs tend to be more robust when considering relative change in FID score.

\section{Disentanglement}
\label{sec:celeba}

The term ``disentanglement'' can cover a broad range of definitions, but a generalized high-level notion is that the model should capture individual factors of variation within linear subspaces of $\mathcal{Z}$. A common goal in the disentanglement literature is to regularize the model in such a way that it automatically aligns factors of variation along the axes of the latent space in an unsupervised manner \cite{Matthey2017betaVAELB,2018arXiv180205983K}. Achieving this goal would therefore mean it is possible to perform semantic manipulation along an individual factor of variation by interpolating along any given axis in $\mathcal{Z}$. Clearly an unregularized autoencoder cannot achieve the same outcome, as it has no incentive to align factors of variation with the axes of $\mathcal{Z}$, and so extra processing is required in order to discover the directions along which factors of variation lie. Despite this lack of automatic discovery, we argue that an unregularized $q(\vect{z})$ with a learned $p(\vect{z})$ has numerous benefits over existing disentanglements methods, which we discuss in the next two subsections.

\subsection{Linear Disentanglement}
Although an unregularized $q(\vect{z})$ does not allow automatic discovery of the factors of variation, we argue that it can achieve greater overall \textit{linear disentanglement}. This means that, for any given semantic attribute, it should be easier to find a linear hyperplane that separates the latent codes into two sets, with each side of the hyperplane corresponding to one of the two possible values of the given attribute. Intuitively, a highly entangled latent representation cannot achieve good linear disentanglement, as it should not be possible to find a linear path in the latent space that can be interpolated along without altering many factors of variation simultaneously, and so semantic attributes cannot lie cleanly on either side of a linear hyperplane. As pointed out in \cite{KarrasStyleGAN2018}, a fixed prior such as a standard normal necessarily entangles the latent space if there is any correlation between factors of variation. In \cite{KarrasStyleGAN2018} the authors posit that the decoder is likely to pressure $q(\vect{z})$ to take on a disentangled form, since intuitively this should make accurate reconstruction easier as opposed to trying to unwarp a highly entangled representation. Therefore it is reconstruction loss, not regularization, that induces better linear disentanglement.

We also argue that an unregularized $q(\vect{z})$ can achieve improved image quality \textit{and} improved disentanglement simultaneously. Following from the point made above, reducing or removing regularization can only be beneficial to disentanglement, as it is the reconstruction loss that induces a disentangled representation. And, given that we have shown in Section \ref{sec:mnist_cifar10} that removing regularization is beneficial to image quality, the two outcomes can be achieved simultaneously.

The CelebA dataset contains 40 binary attributes that we can consider as factors of variation, and thus we can use these attributes to calculate a measure of disentanglement. We consider the linear separability score proposed in \cite{KarrasStyleGAN2018}. In their work they first train a deep network classifier that predicts image attributes on the training images, and then train a linear SVM classifier that predicts the classifier network's output given the latent variable. After this they calculate the conditional entropy $\operatorname{H}(\vect{Y}|\vect{X})$ where $\vect{Y}$ represents the labels predicted by the deep network classifier, and $\vect{X}$ represents the labels predicted by the SVM. It can be seen that lower conditional entropy will correspond to better linear separation, since the SVM will have higher prediction accuracy and thus observing $\vect{Y}$ will give less information. By following their procedure exactly, we can quantitatively measure linear disentanglement purely as a function of the generative process of the model.
We additionally consider the perceptual path length proposed in \cite{KarrasStyleGAN2018}. This gives us a further measure of disentanglement; if the factors of variation lie along paths that are highly warped and curved, then a small movement along a linear (or spherical) path between two randomly sampled endpoints is likely to cause a larger perceptual change than if the factors of variation were lying along linear paths. \\

\begin{figure}[t]
\centering
\begin{subfigure}{0.48\textwidth}
        \includegraphics[width=\textwidth]{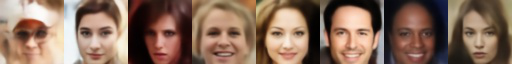} \\
        \includegraphics[width=\textwidth]{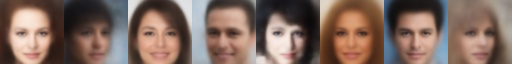} \\
        \includegraphics[width=\textwidth]{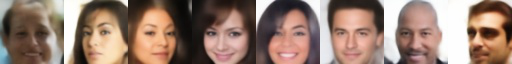} \\
        \includegraphics[width=\textwidth]{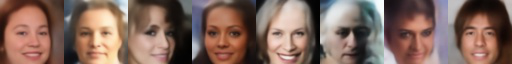} \\
        \includegraphics[width=\textwidth]{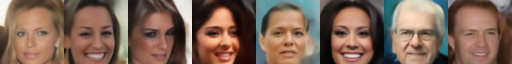}
\end{subfigure}
\begin{subfigure}{0.48\textwidth}
        \includegraphics[width=\textwidth]{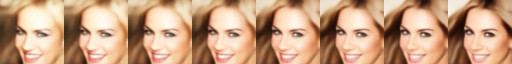} \\
        \includegraphics[width=\textwidth]{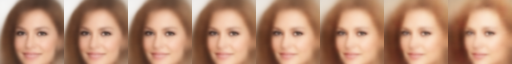} \\
        \includegraphics[width=\textwidth]{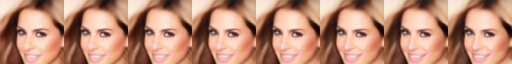} \\
        \includegraphics[width=\textwidth]{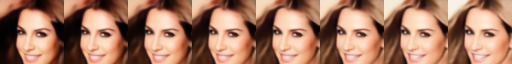} \\
        \includegraphics[width=\textwidth]{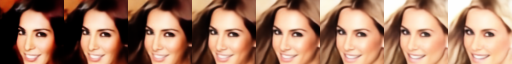}
\end{subfigure}
\caption{Left columns: random samples from each model. Right columns: PCA was performed on $q(z)$ and an image from the dataset was encoded and transformed to PC space. The first principal component was interpolated along $[-1.5,1.5]$ while other components were kept fixed. Row 1: VAE (std. normal prior). Row 2: $\beta$-VAE. Row 3: FactorVAE. Row 4: VAE (flow prior, $\beta = 1$). Row 5: VAE (flow prior, $\beta = 0$).}
\label{fig:celeba_samples_disentanglement}
\end{figure}

We evaluated disentanglement in a VAE with standard normal prior with $\beta = 1$, a $\beta$-VAE \cite{Matthey2017betaVAELB} with $\beta = 25$, a FactorVAE \cite{2018arXiv180205983K} with the total correlation strength set to $40$, a VAE with normalizing flow prior with $\beta = 1$ and a VAE with normalizing flow prior with $\beta = 0$. For both VAEs with normalizing flow prior we evaluated disentanglement of both $\vect{z}_0$ and $\vect{z}_T$ to demonstrate that the fixed base distribution causes the latents to become significantly warped and entangled.
Results are reported in Table \ref{tab:celeba_scores}. The results indicate that when $q(\vect{z})$ is unregularized and not fit to a fixed prior, linear disentanglement is improved, and additionally sample image quality improves. We provide samples from each model in Figure \ref{fig:celeba_samples_disentanglement} as well as an example of feature discovery by performing PCA on $q(\vect{z})$. Clearly PCA is a poor method for feature discovery, however we believe it suffices for this simple demonstration.

\begin{figure}
\centering
\begin{subfigure}{0.9\textwidth}
        \includegraphics[width=\textwidth]{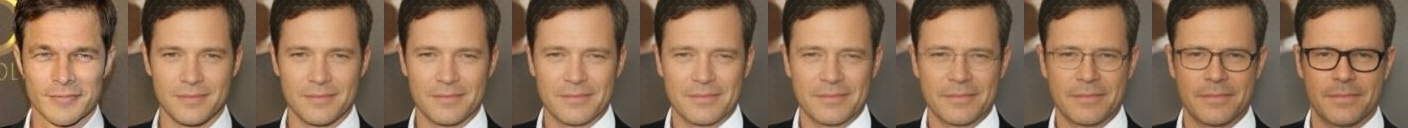}
        \includegraphics[width=\textwidth]{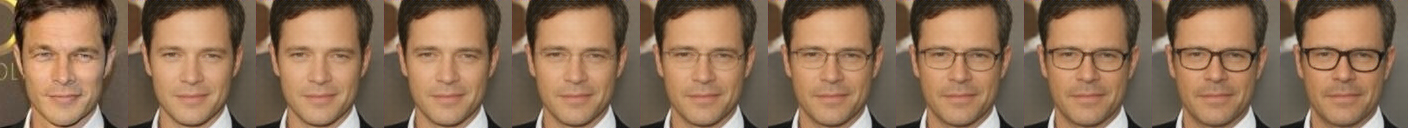} \\
\end{subfigure} \\
\begin{subfigure}{0.9\textwidth}
        \includegraphics[width=\textwidth]{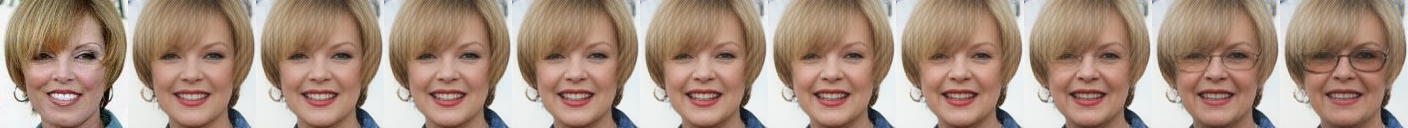}
        \includegraphics[width=\textwidth]{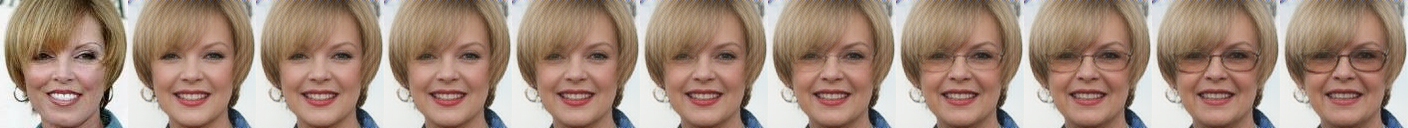}
\end{subfigure}
\caption{Adding glasses to a face. Top row of each set: interpolation in $\mathcal{Z}_0$. Bottom row of each set: interpolation in $\mathcal{Z}_T$. Leftmost column: the original image before encoding. The top row shows a more abrupt change towards the end, while the bottom row is closer to a constant rate of change.}
\label{fig:celeba_semantic_interpolation}
\end{figure}

\subsection{Interpolation}
When interpolating between points in latent space the motivation is often to achieve some semantic mixture between two images, or to change some semantic feature of an image such as putting glasses on a person's face. As discussed in the previous section, poor linear disentanglement is clearly detrimental to this task, as interpolation along a warped latent representation makes it difficult to manipulate a single semantic feature without potentially altering several other unrelated features. Issues of entanglement aside, we expect an unregularized $q(\vect{z})$ will achieve a more constant rate of change when interpolating along the direction of a factor of variation, especially in the case where the factor of variation corresponds to a semantic attribute with class imbalance. This is because when using a fixed prior the rate of change in density is fixed according to the chosen prior distribution, and so the measure of variation is highly unlikely to have a linear correlation with distance travelled along the path. As an example, consider the case of a uniform prior distribution: the relative amount of space occupied by points with a particular semantic feature would be proportional to the class probability of the feature. What this means in practice is that if we were to generate a sequence of images by interpolating along the direction of a semantic attribute with heavy class imbalance, there would be very little change for most of the sequence followed by an abrupt change at the end.

\begin{figure*}[t]
   \centering
\begin{tabular}{>{\centering\arraybackslash}m{0.5cm}ccc}
$\mathcal{Z}_0$&
\raisebox{-.5\height}{\includegraphics[width=0.3\linewidth]{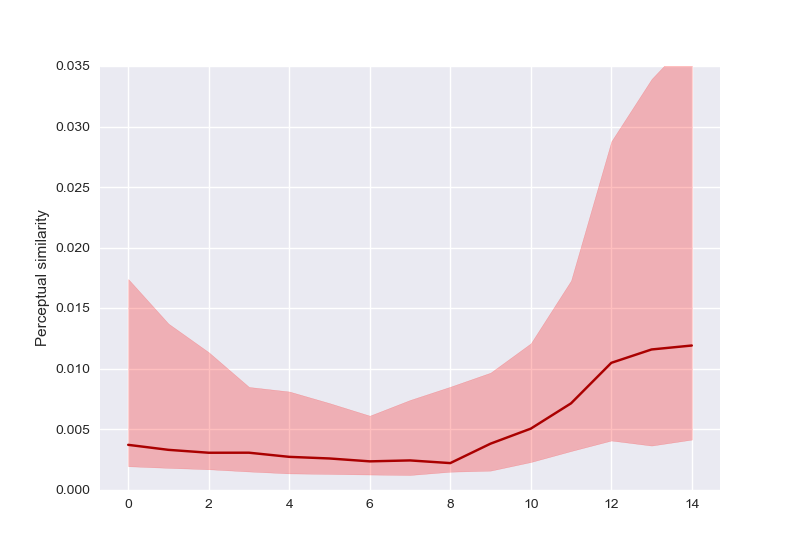}}&
\raisebox{-.5\height}{\includegraphics[width=0.3\linewidth]{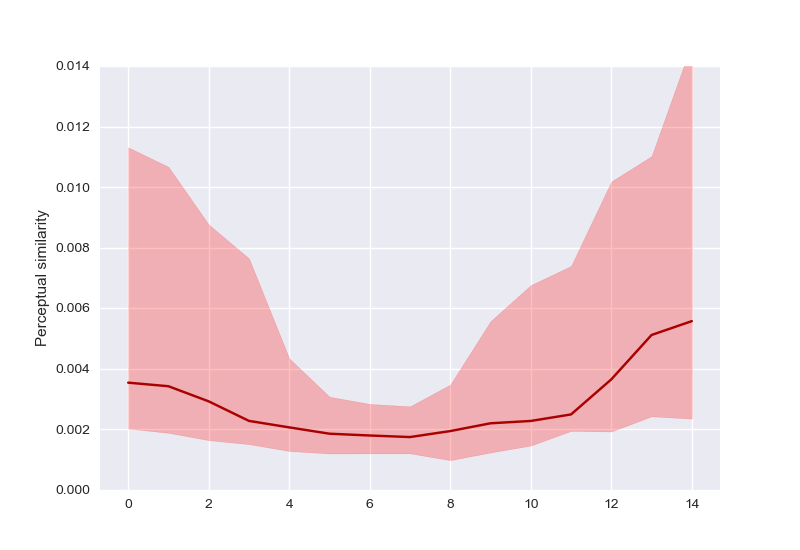}}&
\raisebox{-.5\height}{\includegraphics[width=0.3\linewidth]{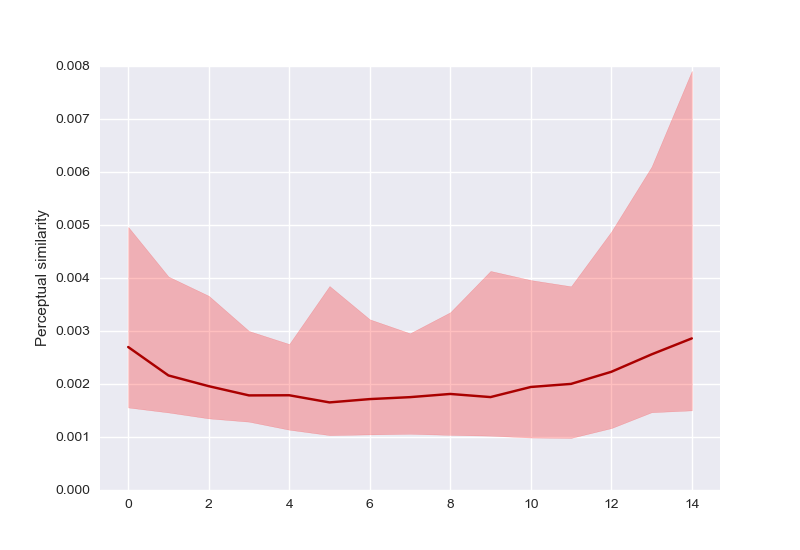}}\\
$\mathcal{Z}_T$&
\raisebox{-.5\height}{\includegraphics[width=0.3\linewidth]{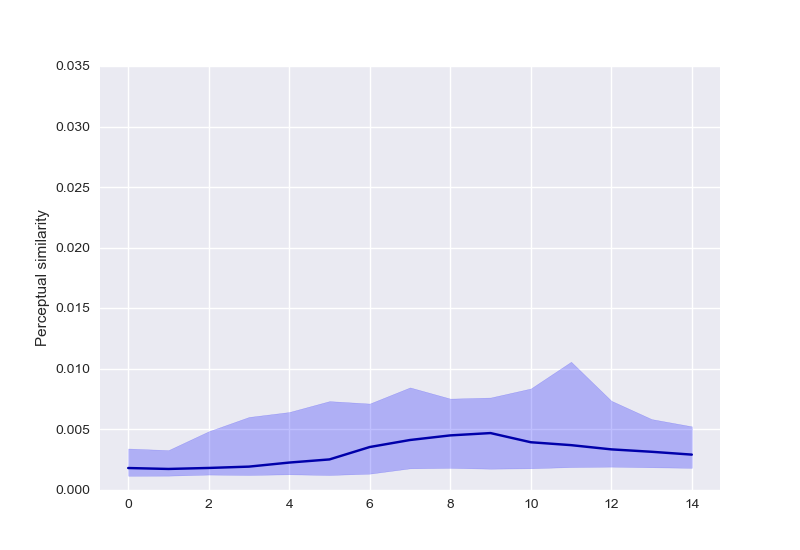}}&
\raisebox{-.5\height}{\includegraphics[width=0.3\linewidth]{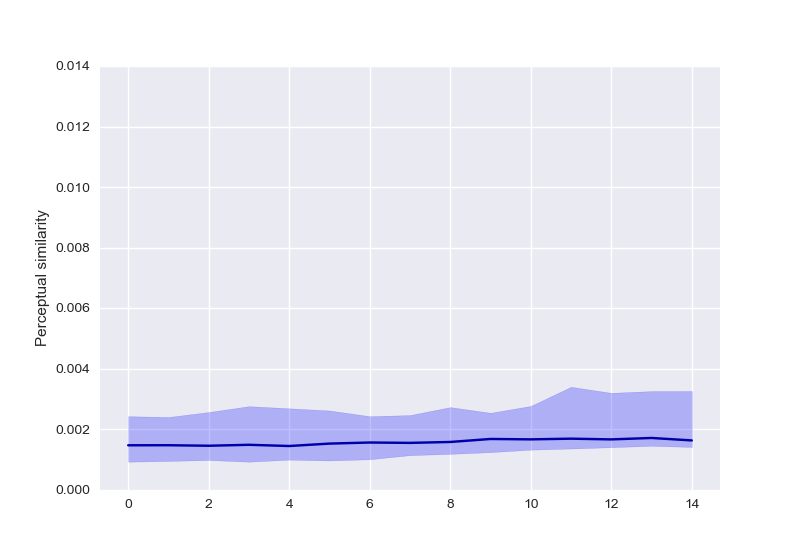}}&
\raisebox{-.5\height}{\includegraphics[width=0.3\linewidth]{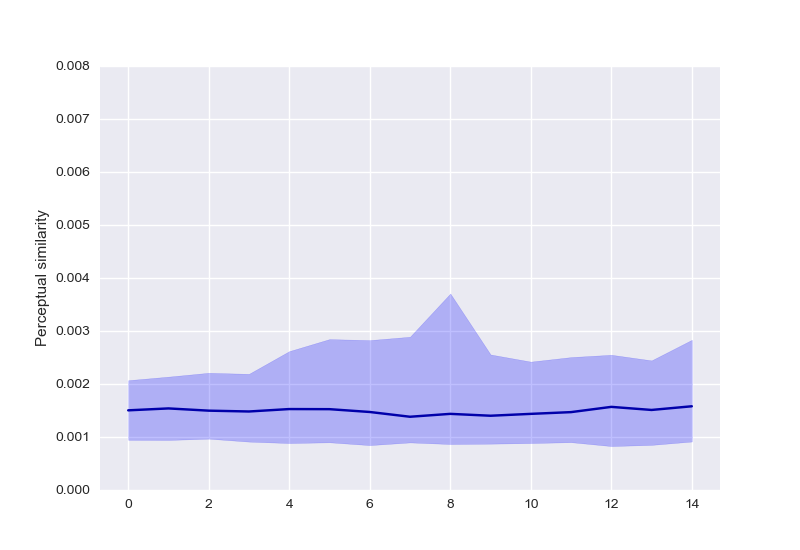}}\\
&Glasses&Beard&Blond\\
\end{tabular}
    \caption{We plot the median perceptual change for each point in multiple sequences generated by manipulating a particular semantic feature in either $\mathcal{Z}_0$ or $\mathcal{Z}_T$. The shaded region represents the 5th to 95th percentiles. X-axis represents index in the sequence. Y-axis represents perceptual change.}
    \label{fig:interpolation_quantitative}
\end{figure*}

A normalizing flow prior allows us to perform a fair comparison between interpolation in an unregularized distribution and interpolation in a fixed distribution, since the the bijection allows us to interpolate between corresponding path endpoints in either distribution. We therefore experiment on the difference between interpolating in $\mathcal{Z}_0$ or in $\mathcal{Z}_T$ using a VAE with normalizing flow prior and $\beta = 0$. As a point of clarification, when we say we are interpolating between two images $\vect{x}^{(a)}$ and $\vect{x}^{(b)}$ in $\mathcal{Z}_T$ we are calculating $\hat{\vect{x}} \sim p(\vect{x}|\hat{\vect{z}})$ where $\hat{\vect{z}} = \operatorname{lerp}(\vect{z}^{(a)}, \vect{z}^{(b)}; t)$, $\vect{z}^{(a)} \sim q(\vect{z}|\vect{x}^{(a)})$, $\vect{z}^{(b)} \sim q(\vect{z}|\vect{x}^{(b)})$ and $t$ varies between $0$ and $1$. When we are interpolating in $\mathcal{Z}_0$ we are instead calculating $\hat{\vect{z}} = f^{-1}(\operatorname{slerp}(f(\vect{z}^{(a)}), f(\vect{z}^{(b)}); t))$, where $f$ is the bijection defined by the normalizing flow.
In order to calculate the direction of change for a particular attribute, we first calculate the mean of all latent codes $\vect{z}_T$ corresponding to images with and without the attribute. We then calculate the difference between these two means to produce the direction of change. If the attribute corresponds to glasses, for example, we can encode an image of a person not wearing glasses, add the vector representing the direction of change to the latent encoding, and then decode to produce an image of the same person wearing glasses. We used VGG19 perceptual loss for the decoder of the model as we found this helped with semantic manipulation using vector arithmetic. To quantitatively measure the rate of change, we sample 5000 images from the data set without the attribute, and then interpolate evenly along the direction of change to produce a sequence of 16 images. We then measure perceptual difference between adjacent images using a VGG16 network. Results are show in Figure \ref{fig:interpolation_quantitative}, where we plot the median perceptual change along the generated sequences. It can be seen that the rate of change when interpolating in $\mathcal{Z}_0$ varies significantly, especially at the ends of the sequence. Interpolation in $\mathcal{Z}_T$ on the other hand produces a rate of change that is much closer to constant, which is more ideal for semantic manipulation. Some sample sequences are shown in Figure \ref{fig:celeba_semantic_interpolation}.

For architecture and training details regarding our disentanglement experiments, please refer to the supplementary.

{\setlength{\tabcolsep}{1.5em}
\begin{table}
\centering
\captionof{table}{Linear separability, perceptual path length (PPL) and FID scores after training on CelebA.}
    \begin{tabular}{| l l l l |}
    \hline
     & Separability & PPL & FID \\ \hline
    VAE (std. normal prior) & $2.14$ & $1139$ & $41.4$ \\ \hline
    $\beta$-VAE & $2.09$ & $1324$ & $82.7$ \\ \hline
    FactorVAE & 2.32 & $1339$ & $39.8$ \\ \hline
    VAE (flow prior, $\beta = 1$) ($\vect{z_0}$) & $2.46$ & $1753$ & \multirow{2}{*}{$38.7$} \\
    VAE (flow prior, $\beta = 1$) ($\vect{z_T}$) & $1.68$ & $1173$ & \\ \hline
    VAE (flow prior, $\beta = 0$) ($\vect{z_0}$) & $2.82$ & $1933$ & \multirow{2}{*}{$\bm{33.1}$} \\
    VAE (flow prior, $\beta = 0$) ($\vect{z_T}$) & $\bm{1.67}$ & $\bm{1076}$ & \\
    \hline
    \end{tabular}
\label{tab:celeba_scores}
\end{table}}

{\setlength{\tabcolsep}{1.5em}
\begin{table}
\centering
\captionof{table}{Inception and LPIPS scores (higher is better) on the DeepFashion dataset after training Variational U-net.} \label{tab:deepfashion} 
    \begin{tabular}{| l l l |}
    \hline
     & IS & LPIPS \\ \hline
    VUNET (Original) & $2.63$ & $0.184$ \\
    VUNET (Proposed) & $2.70$ & $0.236$ \\
    \hline
    \end{tabular}
\label{tab:vunet_scores}
\end{table}}

\begin{figure}[h]
\centering
\begin{tabular}{r l}
        \includegraphics[height=1.25cm]{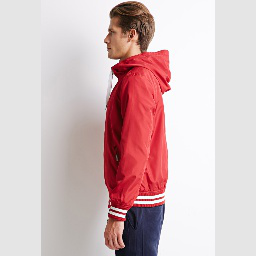} &
        \includegraphics[height=1.25cm]{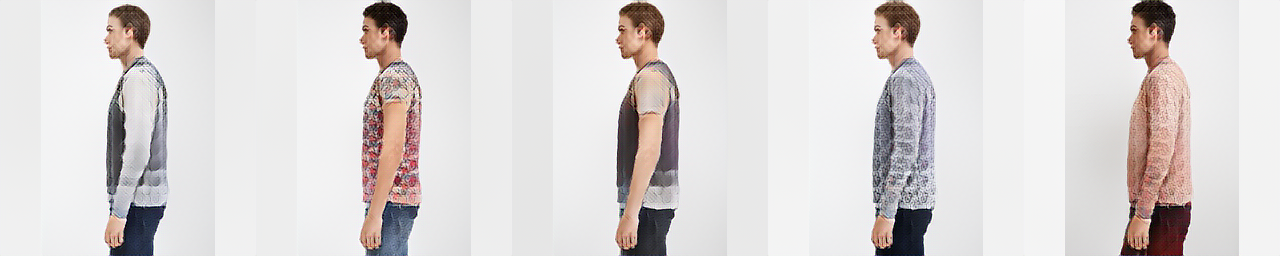} \\
        \includegraphics[height=1.25cm]{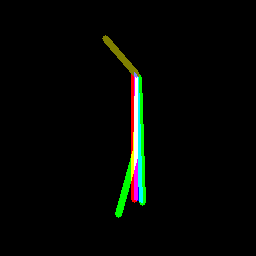} &
        \includegraphics[height=1.25cm]{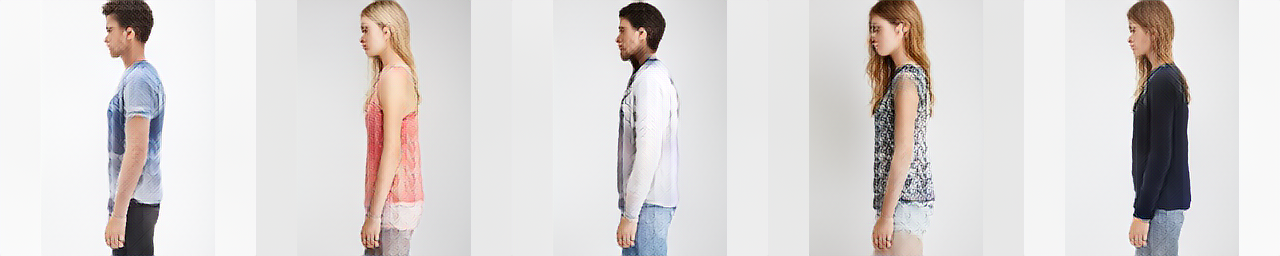} \\
        \addlinespace
        \includegraphics[height=1.25cm]{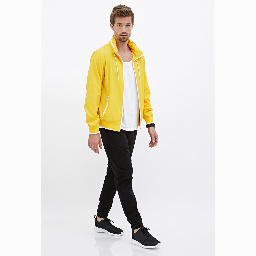} &
        \includegraphics[height=1.25cm]{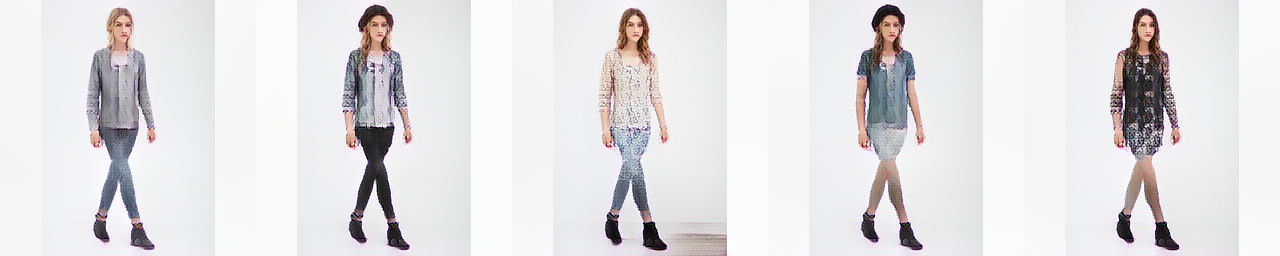} \\
        \includegraphics[height=1.25cm]{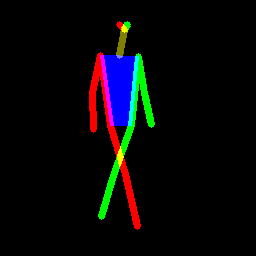} &
        \includegraphics[height=1.25cm]{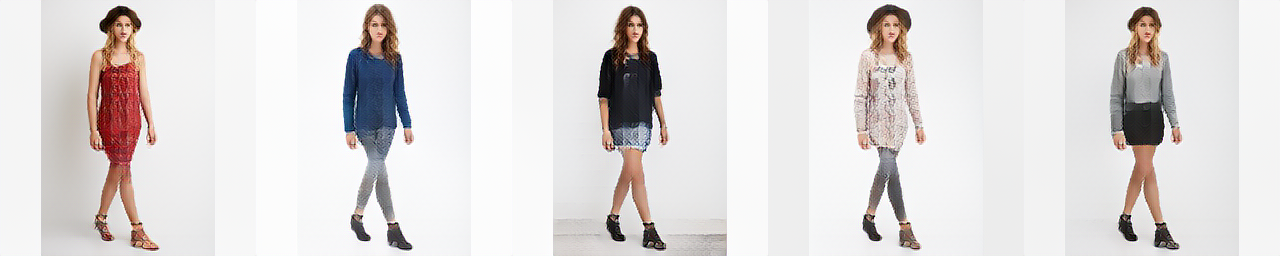} \\
        \addlinespace
        \includegraphics[height=1.25cm]{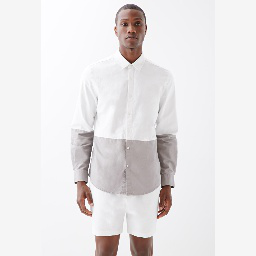} &
        \includegraphics[height=1.25cm]{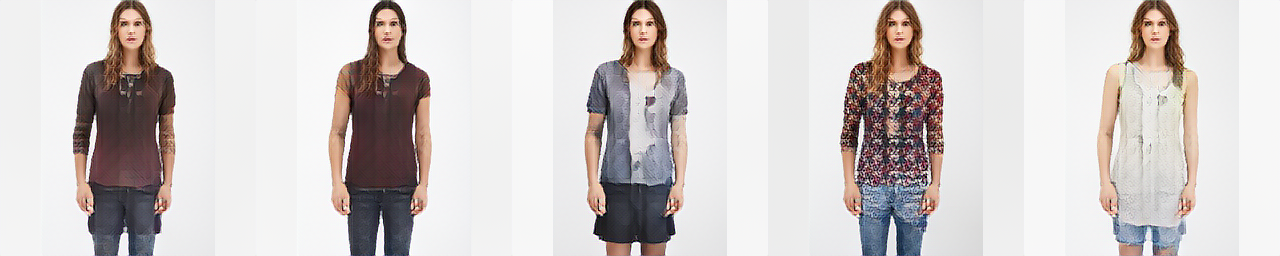} \\
        \includegraphics[height=1.25cm]{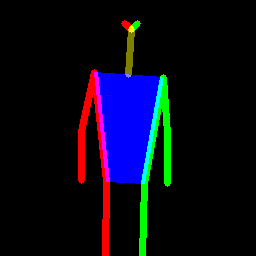} &
        \includegraphics[height=1.25cm]{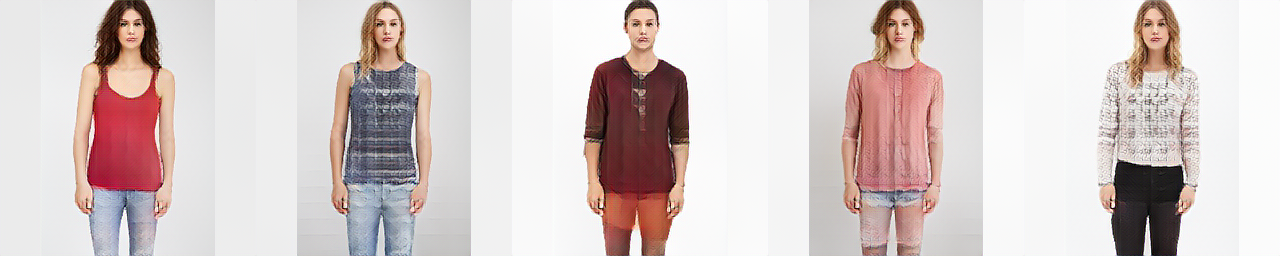}
        \label{fig:vunet3}
\end{tabular}
\caption{Conditional samples using Variational U-net. For each set, the top row contains samples from the original, while the bottom row contains samples from the proposed change. Leftmost column contains the original image and the pose being conditioned on.}
\label{fig:vunet_samples}
\end{figure}

\section{Diversity in image-to-image translation}
\label{sec:vunet}
Achieving high sample diversity in multi-modal image-to-image translation tasks is often an explicit goal \cite{ZhuBicycleGAN2017,2018_MUNIT_Huang}. When using conditional VAEs for image-to-image translation tasks, the decoder is often able to learn a fairly accurate reconstruction based on the conditioned image alone, and so may ignore the latent code entirely if the KL divergence weight is too strong. In order to quantitatively test whether our proposed method is able improve diversity, we experiment with the Variational U-net model proposed in \cite{Esser_2018_CVPR}. In their work, they attempt to learn the distribution over images of people conditioned on their pose. In their implementation they make use of KL divergence annealing in order to encourage the model to make use of the latents, however sample diversity may still be negatively affected. We modified their implementation such that the prior distribution is learned via normalizing flow, and dropped the KL divergence term from the objective. Their model conditions the prior distribution on the given pose such that their objective becomes
\begin{equation} \log p(\vect{x}^{(i)}|\vect{y}^{(i)},\vect{z}) - D_{KL}[q(\vect{z}|\vect{x}^{(i)},\vect{y}^{(i)})||p(\vect{z}|\vect{y}^{(i)})] \end{equation}
where $\vect{y}$ is the pose and $\vect{x}$ is the real image. For compatibility we therefore learn the mean of $\vect{z_0}$ conditioned on the pose and additionally condition each flow transformation $f_t$ on the pose. To measure diversity, we compute the LPIPS distance \cite{Zhang2018perceptual} between randomly sampled pairs which were generated by conditioning on the images in the test set. We additionally calculated the Inception score \cite{ImprovedGAN2016} of the samples to ensure image quality was not affected. Results comparing our modification with the original implementation after training on the DeepFashion dataset are reported in Table \ref{tab:deepfashion}. We also show samples in Figure \ref{fig:vunet_samples}.

\section{Conclusion}
We have proposed removing latent regularization from the objective of generative autoencoder models in the case where the prior is sufficiently expressive. We demonstrated empirically that this results in improved image quality, improved linear disentanglement, and improved sample diversity. Our results indicate that fixed-form priors should be eschewed in favour of learned priors with little to no latent regularization.

%
%
\bibliographystyle{splncs04}
\bibliography{egbib}

\appendix
\section{Regularization experiment}
We use the same autoencoder architecture used in \cite{ChenDHSSA16} for all datasets.

\subsection{Flow prior architecture}
We use the same normalizing flow architecture to model the prior in each of our experiments. Here we give a detailed overview of the architecture.
The type of flow we used is the same as that of RealNVP \cite{DinhSB16}. That is, for each individual transformation in the flow we split the input into two parts, and pass one part through a neural network to give the parameters of an affine transformation that is applied to the second part. We additionally apply an affine transformation with learnable parameters before every second transformation, the same as the \texttt{actnorm} operation used in \cite{KingmaGlow}. We do not use data-dependent initialization for the affine transformation, and instead initialize it as the identity function. Pseudo-code for a single transformation $f_t$ within the flow is given below: \\


\begin{figure}[ht]
  \centering
  \begin{minipage}{.9\linewidth}
\begin{algorithmic}
\State {\jingAlgo{z} $\gamma z + \beta$} \Comment{affine transformation}
\State {\jingAlgo{z_a,z_b} $\operatorname{split}(z)$}
\State {\jingAlgo{\mu,\sigma} $\operatorname{NN}(z_a)$}
\State {\jingAlgo{\sigma} $\operatorname{sigmoid}(\sigma + 2)$}
\State {\jingAlgo{z_b} $\sigma(z_b + \mu)$}
\State {\jingAlgo{z} $\operatorname{concat}(z_a, z_b)$}
\end{algorithmic}
  \end{minipage}
\end{figure}

Next we discuss the structure of the network \texttt{NN}. We use the following denotations to describe the architecture: \texttt{FC-X} denotes a fully connected layer with an \texttt{X}-dimensional output. Let $D$ donate the ``width'' of a flow transform, and let $R$ equal $\dim(\mathcal{Z}) / 2$. Then \texttt{NN} is given by \texttt{FC-D $\rightarrow$ ReLU $\rightarrow$ FC-D $\rightarrow$ Relu $\rightarrow$ FC-R}. \\

All kernel weights in \texttt{NN} are initialized using Glorot uniform initialization \cite{pmlr-v9-glorot10a}, except for the final fully connected layer whose weights are initialized as zero. \\

\subsection{Adversarial prior architecture}

The adversarial prior architecture used across all experiments is as follows. Gaussian noise is passed through several layers of width $D$, followed by a final fully connected layer of width $R$. Let \texttt{BN} denote batch normalization. Each layer is given by \texttt{FC-D $\rightarrow$ BN $\rightarrow$ ReLU}. \\

The discriminator is also given by several layers of width $D$ followed by a final fully connected layer of width $1$. Each layer is given by \texttt{FC-D $\rightarrow$ ReLU}.

\subsection{Hyperparameters and training}
A batch size of 100 and an initial learning rate of 0.0001 was used across all datasets and models. We used the Adam optimizer \cite{2014arXiv1412.6980K} with default parameters. \\

The details for the flow prior experiments are as follows. \\ For MNIST we used a 24 layer flow with a width of 1024. The entire model was trained for 200 epochs, and the prior was then trained independently for a further 100 epochs. \\ For Fashion-MNIST we used a 24 layer flow with a width of 1024. The entire model was trained for 300 epochs, and the prior was then trained independently for a further 100 epochs. \\ For CIFAR-10 we used a 16 layer flow with a width of 256. The entire model was trained for 500 epochs, and the prior was then trained independently for a further 100 epochs. The learning rate was halved every 250 epochs. \\ For CelebA we used a 30 layer flow with a width of 1024. The entire model was trained for 200 epochs, and the prior was then trained independently for a further 50 epochs. \\

The details for the adversarial prior experiments are as follows. We used the same size prior and discriminator across all datasets; the prior was 3 layers of width 1024 followed by a final fully connected layer and the discriminator was 2 layers of width 1024 followed by a final fully connected layer. \\ For MNIST we trained the entire model for 200 epochs. \\ For Fashion-MNIST we trained the entire model for 300 epochs. \\ For CIFAR-10 we trained the entire model for 300 epochs. \\ For CelebA we trained the entire model for 200 epochs.

\section{Disentanglement experiments}
For our disentanglement experiments using CelebA we used exactly the same Resnet architecture as described in \cite{Dai2018diagnosing} with a depth of 4. The prior architecture used for the VAEs with normalizing flow prior was the same as described in the previous section. L2 loss was used for the decoder. We trained for 300 epochs and halved the learning rate every 100 epochs.
\subsection{VGG19 for interpolation}
We used the squared difference between the hidden features of the \texttt{relu\_1\_1}, \texttt{relu\_2\_1}, \texttt{relu\_3\_1} and
\texttt{relu\_4\_1} layers of the VGG19 network as the reconstruction loss for the interpolation experiment.

\end{document}